# Interpolated Discretized Embedding of Single Vectors and Vector Pairs for Classification, Metric Learning and Distance Approximation

Ofir Pele and Yakir Ben-Aliz

*Center for Cyber Technology,*
*Department of Computer Science,*
*Department of Electrical and Electronics Engineering,*
*Ariel University*

August 8, 2016

**Abstract**

We propose a new embedding method for a single vector and for a pair of vectors. This embedding method enables: *a*) efficient classification and regression of functions of single vectors; *b*) efficient approximation of distance functions; and *c*) general, non-Euclidean, semimetric learning. To the best of our knowledge, this is the first work that enables learning any general, non-Euclidean, semimetrics. That is, our method is a universal semimetric learning and approximation method that can approximate any distance function with as high accuracy as needed with or without semimetric constraints. The project homepage including code is at: `http://www.ariel.ac.il/sites/ofirpele/ID/`

## 1 Introduction

Distance functions are at the core of numerous scientific areas. One can choose a distance function based on prior knowledge or learn it from data, *metric learning*. The most commonly used distance function is the Euclidean distance. In metric learning, most of the works learn a Mahalanobis distance. These methods [1–12] learn a linear transform that is applied on each vector and then apply the squared Euclidean distance (thus these methods are actually semimetric learning). Kernel metric learning applies embedding separably on each vector before learning the linear transform (the embedding is done usually implicitly using the kernel trick). Deep learning methods (such as [13]) learn an embedding using a deep network and then apply the Euclidean distance on the embedded vectors (the output of the network). Thus, even kernel and deep metric learning can only learn a Euclidean distance. Some works [14–17] have suggested learning other families of distances. However, these methods are restricted to the suggested pre-chosen families of distances (*e.g.* Earth Mover's Distance [14, 15] and $\chi^2$ [16, 17]). Additionally, they are non-convex. Finally, multi-metric learning methods [1, 18] learn separate local Mahalanobis metrics around keypoints. However, they do not learn a global metric. An exception is [19] which shows how to combine information from several local metrics into one global metric. However, again it is only able to model Euclidean metrics.

In this paper, we present a method for the approximation and learning of arbitrary continuous distance functions and semimetrics. That is, a universal semimetric learning and approximation method. The main difference between our model and previous models is that our model embeds a pair of vectors *jointly* and not separably. The distance between the vectors is the embedded vector dot product with a vector of parameters that can be learned. Thus, many of the learning objectives are convex in our model. Additionally, we can enforce constraints on the vector of parameters such that the resulting distance will be a continuous semimetric. To the best of our knowledge, this is the first work that enables learning and approximation of arbitrary distance functions or arbitrary semimetrics (where approximation is done by setting the parameters instead of learning them).

Our method is based on discretization and interpolation of $n$-dimensional vector. The discretization allows us to model different regions of the features space separately, while the interpolation preserves the original continuous values.

Our embedding can also be applied to single vectors as an efficient non-linear learning method as was done by [20–25]. Our embedding method embeds a vector into a high dimensional but sparse vector. Thus, learning a linear classifier in this representation is efficient.

In the following we will describe the most relevant previous works. Maji et al. [21] approximate any additive kernel using single feature discretization and linear interpolation (although the motivation was to approximate the histogram intersection kernel). This enables learning an approximation of any additive model with a time complexity that is linear in the dimension of the input. Pele et al. [20] added second order terms that approximate cross-relationships between two features. Pele et al. [26] showed that the same embedding enables learning any bin-to-bin distance. Bin-to-bin distances are distances that compare corresponding bins of a vector to its exact corresponding bin in the second vector (*e.g.* the $i$th entry of $\overrightarrow{x_1}$ is compared only to the $i$th entry of $\overrightarrow{x_2}$). Bernal et al. [27] suggested adding $n$-features relationships for a single vector, called "Piecewise-N-linear binning". This method uses a different interpolation scheme that like our method its memory complexity is exponential in the number of features used, but in contrary to our method its time complexity is also exponential. Whereas, in our case the time complexity is only $n \lg nc$. Additionally, our method enables modeling general, non-Euclidean, semimetric distance functions.

**Algorithm 1** ID embedding algorithm
___
**Input:** $\vec{x} \in \mathbb{R}^n$
**Input:** $V \in \mathbb{R}^{n \times c}$ a matrix that contains sorted discretization points for each dimension in each row (see Eq. 1).
**Output:** $\vec{\phi} \in [0,1]^{c^n}$ a sparse vector
1: $\phi$ = empty (all zeros), $c^n$-dimensional, sparse vector
2: \\ Clips to the maximum or minimum respectively and then finds the
3: \\ largest per-coordinate vertex of the hyperrectangle in which
4: \\ the clipped point $\vec{x}$ is in and normalize to unit hyperrectangle
5: **for** $i = 0 \ldots n-1$ **do**
6: $\quad x_i = \min(\max(x_i, v_{i,0}), v_{i,c-1})$
7: $\quad d_i = \min_{j \in [1,2,\ldots,c-1]} s.t : x_i \leq v_{i,j}$
8: $\quad \hat{x}_i = \frac{x_i - v_{i,d_i-1}}{v_{i,d_i} - v_{i,d_i-1}}$
9: \\ Finds the simplex (permutation)
10: $[s_0, s_1, \ldots, s_{n-1}]$ = sort $[0, 1, \ldots, n-1]$ according to $[\hat{x}_0, \hat{x}_1, \ldots, \hat{x}_{n-1}]$
11: \\ Computes $n+1$ barycentric coordinates of $\vec{x}$ in the simplex
12: $p = \sum_{i=0}^{n-1} d_i c^i$
13: $\sigma = 0$
14: **for** $i = 0 \ldots n-1$ **do**
15: $\quad \phi_p = \hat{x}_{s_i} - \sigma$
16: $\quad \sigma = \sigma + \phi_p$
17: $\quad p = p - c^{s_i}$
18: $\phi_p = 1 - \sigma$
19: **return** $\vec{\phi}$
___

## 2 Interpolated Discretized Embedding (ID) for a Single Vector

In this section we describe the ID embedding function. The input for ID is an $n$-dimensional vector: $\vec{x} \in \mathbb{R}^n$. ID is parameterized by an $n \times c$ matrix that contains sorted discretization points for each dimension in each row (the generalization to a different number of discretization points for each dimension is trivial):

$$V = \begin{bmatrix} v_{0,0} & v_{0,1} & \cdots & v_{0,c-1} \\ \vdots & \vdots & \ddots & \vdots \\ v_{n-1,0} & v_{n-1,1} & \cdots & v_{n-1,c-1} \end{bmatrix} \quad (1)$$

$$\forall \ 0 \leq i < n, \quad v_{i,0} \leq v_{i,1} \leq \cdots \leq v_{i,c-1}$$

The output of ID is a sparse vector (at most $n+1$ non-zero values) in the $c^n$-unit-hypercube: $\text{ID}(\vec{x}; V) \in [0,1]^{c^n}$. The algorithm is described in Alg. 1. The algorithm first initializes an empty (all zeros), $c^n$-dimensional, sparse vector which will contain at most $n+1$ non-zeros entries when it is returned at the end of the algorithm. The time complexity of this step is $O(1)$.

In lines 5-8 the algorithm clips to the maximum or minimum respectively and then finds the largest per-coordinate vertex of the hyperrectangle in which the clipped point $\vec{x}$ is in (in the rectangular grid in the $n$-dimensional space that is determined by $V$). Additionally, the algorithm *implicitly* normalizes the $n$-hyperrectangle into a $n$-unit-hypercube (where the largest per-coordinate vertex of the hypercube is the $n$-dimensional vector: $[1, \ldots, 1]$). The point $\vec{x} = [x_1, \ldots, x_n]$ is *explicitly* normalized into $\vec{\hat{x}} = [\hat{x}_0, \ldots, \hat{x}_{n-1}]$. The time complexity of this step is $O(n \lg c)$ (as $V$ rows are sorted).

The next step is to find out in which simplex the concatenated vector is. We dissect each hypercube into $n!$ simplexes according to its permutation [28]. In line 10 we find the simplex in which $\vec{\hat{x}}$ is. The time complexity of this step is $O(n \lg n)$.

The vertices of the simplex are the vertices of the unit-hypercube that have the same permutation as $\vec{\hat{x}}$. Thus, they are the $n+1$ points:

$$[0, \ldots, 0], [0, \ldots, \underbrace{1}_{s_{n-1}}, \ldots, 0], \ldots, [1, \ldots, \underbrace{0}_{s_0}, \ldots, 1], [1, \ldots, 1] \qquad (2)$$

To find the barycentric coordinates $\lambda_0, \ldots, \lambda_n$ of $\vec{\hat{x}}$, we need to solve this set of $n+1$ equations with $n+1$ variables:

$$[0, \ldots, 0]\lambda_0 + [0, \ldots, \underbrace{1}_{s_{n-1}}, \ldots, 0]\lambda_1 + \cdots + [1, \ldots, \underbrace{0}_{s_0}, \ldots, 1]\lambda_{n-1} + [1, \ldots, 1]\lambda_n = [\hat{x}_0, \ldots, \hat{x}_{n-1}]$$
$$\sum_{i=0}^{n} \lambda_i = 1$$
$$(3)$$

Lines 12-18 computes these barycentric coordinates efficiently and stores them in the correct index in $\phi$. We exploit here the fact that we can compute the vertex $[d_0 + 1, \ldots, d_{n-1} + 1]$ index and then update it for all other vertices by reducing the right power of $c$. As the powers of $c$ can be precomputed, the time complexity of this step is $O(n)$.

Clearly, as the number of discretization points increases, we can approximate any Lipschitz continuous function.

The time complexity of ID is $O(n \lg nc)$. Our embedding method embeds a vector into a high dimensional but sparse vector. Thus, learning a linear classifier in this representation is efficient. The memory complexity, however is high: $O(c^n)$. As the embedding is into a sparse vector, the memory that we need for learning from $k$ objects can be reduced to $O(kn)$ and then the time complexity is $O(n \lg nck)$.

A possible regularization that also reduces memory complexity is to use groups of indices. That is, let $\mathcal{S} = \{S_i\}_{i=1}^{g}, S_i \subseteq [1, \ldots, n]$, we define the ID embedding with respect to these groups as the concatenation of embedding each sub-vector (defined by the indices):

$$\text{ID}(\vec{x}; \mathcal{S}) = [\text{ID}(\vec{x}_{S_1}) \ldots \text{ID}(\vec{x}_{S_g})] \tag{4}$$

For example, using groups of size $k$ the time complexity is $O(\binom{n}{k} k \lg ck)$, but using caching and merging it can be reduced to $O(\min(n \lg nc + \binom{n}{k}k, n \lg c + \binom{n}{k} k \lg k))$.

## 3 Interpolated Discretized Embedding (ID) for a Pair of Vectors

Let $[\vec{x_1}, \vec{x_2}] \in [l, u]^{2n}$ be a concatenation of the two vectors and let $V' = [V; V] \in [l, u]^{2n \times c}$ be a concatenation of $V$ (see Eq. 1) to itself row-wise. In this section we show that we can constraint $\vec{w} \in \mathbb{R}^{c^{2n}}$ entries such that $d(\vec{x_1}, \vec{x_2}; \vec{w}) = \text{ID}([\vec{x_1}, \vec{x_2}]) \cdot \vec{w}$ will be a semimetric.

Semimetrics are distance functions between two objects $o_1, o_2$ that satisfy the following properties:

$$d(o_1, o_2) \geq 0 \text{ (non-negativity)} \tag{5}$$
$$d(o_1, o_2) = 0 \text{ if and only if } o_1 = o_2 \text{ (identity of indiscernibles)} \tag{6}$$
$$d(o_1, o_2) = d(o_2, o_1) \text{ (symmetry)} \tag{7}$$

As $\text{ID}([\vec{x_1}, \vec{x_2}]) \in [0, 1]^{c^{2n}}$, if we constraint $\forall \ 0 \leq p < c^{2n}, \ w_p \geq 0$ then $d(\vec{x_1}, \vec{x_2}; \vec{w}) = \text{ID}([\vec{x_1}, \vec{x_2}]) \cdot \vec{w}$ will be non-negative.

For the first side of identity of indiscernibles we need to show that if $\vec{x_1} = \vec{x_2}$ then $d(\vec{x_1}, \vec{x_2}; \vec{w}) = \text{ID}([\vec{x_1}, \vec{x_2}]) \cdot \vec{w} = 0$. For this we need to constraint $\vec{w}$ entries to be zero for the discrete points in the grid where the first vector is equal to the second:

$$\forall \ ((p = \sum_{i=0}^{2n-1} d_i c^i) \wedge (0 \leq d_i < c) \wedge (\forall \ 0 \leq i < 2n-1, \ d_i = d_{n+i})), \ w_p = 0 \tag{8}$$

Eq. 8 is one side of the identity of indiscernibles property for points that are on the discrete grid. The generalization for other points follows from that $\text{ID}([\vec{x_1}, \vec{x_2}])$ contains barycentric coordinates of the simplex in which $[\vec{x_1}, \vec{x_2}]$ is in. If $\vec{x_1} = \vec{x_2}$ then the point $[\vec{x_1}, \vec{x_2}]$ will be in several simplices, where all of them share vertcies that satisfy Eq. 8. Thus, the interpolation will be only from zero-valued vertices and it will also be zero.

For the other side of the identity of indiscernibles we need to show that if $\vec{x_1} \neq \vec{x_2}$ then $d(\vec{x_1}, \vec{x_2}; \vec{w}) = \text{ID}([\vec{x_1}, \vec{x_2}]) \cdot \vec{w} > 0$. This is easily achievable by adding $\varepsilon > 0$ to the distance if the vectors differ before the clipping to maximum and minimum and by constrainting $\vec{w}$ entries to be non zeroes for discrete points in the grid where the first vector is not equal to the second:

$$\forall \; ((p = \sum_{i=0}^{2n-1} d_i c^i) \wedge (0 \leq d_i < c) \wedge (\exists \; 0 \leq i < 2n-1, \; d_i \neq d_{n+i})), \; w_p > 0 \tag{9}$$

Eq. 9 is the other side of the identity of indiscernibles property for points that are on the discrete grid. The generalization for other points follows from the same reasons as the first side.

Finally, we need to show how to achieve symmetry. This is done by a simple change of the ID embedding function. We can add a line that swaps the vectors if the first vector is lexicographically smaller than the second vector.

As we embed the two vectors *jointly*, it is clear that as the number of discretization points increases, we can approximate any Lipschitz continuous distance measure and impose any of the semimetrics constrains. Thus, our method can approximate or learn any semimetric and not just the Euclidean or other families of distances.

As we mentioned in Section 2, the time complexity of ID is $O(n \lg nc)$. Thus, the time complexity of a naïve algorithm for finding the nearest neighbor in a dataset that contains $u$ examples, is $O(un \lg nc)$. Instead, we can clip to the maximum or minimum respectively and then find the largest per-coordinate vertex of the hyperrectangle in which the clipped point is in and normalize to unit hyperrectangle and then find the simplex (permutation) for each point in the dataset in an *offline* stage. In the online stage, we can do the same only for the single point and merge in order to get the full permutation. Thus, finding the nearest neighbor time complexity is $O(n \lg cn + un)$. Similarly, computing all distances between $u$ examples time complexity is $O(un \lg cn + u^2 n)$.

## 4 Applications of Interpolated Discretized Embedding (ID)

There are many possible applications of ID. ID can be applied to single vectors as an efficient non-linear learning method as was done by [20–25]. Our embedding method embeds a vector into a high dimensional but sparse vector. Thus, learning a linear classifier in this representation is efficient.

Many metric learning methods can use $d(\vec{x_1}, \vec{x_2}; \vec{w}) = \text{ID}([\vec{x_1}, \vec{x_2}]) \cdot \vec{w}$ instead of the squared Mahalanobis distance $d(\vec{x_1}, \vec{x_2}; A) = (\vec{x_1} - \vec{x_2})^T A (\vec{x_1} - \vec{x_2})$ and thus be universal distance function learning methods. Learning is often convex as the distance is a linear function. Enforcing semimetric properties can easily be done as these are linear constraints. The one dimensional ID was used by Pele et al. [26] with the Large Margin Metric Nearest Neighbor (LMNN) learning framework [1]. We note that embedding all pairs with ID enables learning any distance that takes pairs of features into account. If $c = O(n)$ then time complexity is $O(n^2)$, just like regular Mahalanobis learning but we can learn $O(n^2 c^2)$ parameters of the distance. Additionally, the distance is not restricted to Euclidean.

It is useful to have an approximation of distances that their computation have high time complexity, such as the Earth Mover's Distance [29–31]. Most of these works include embedding the vectors to a different space and then using a distance such as $L^1$ [32–35]. However, it is well known that not all distances are embeddable to $L^1$ or $L^2$ [36]. Using ID we can approximate any distance as accurate as we want by increasing the number of discretization points. The main difference between our method and previous work is that our method embeds the points *jointly* and not separably.

In *distance regression*, we do not know the distance function and we would like to learn it from examples. That is, we are given examples, where each example is a pair of two vectors and their distance and we would like to learn an approximation of the true *unknown* distance function also for unseen pairs of vectors. There has not been much work on these problems [19, 37]. Using ID metric regression can be easily modeled as a least squares or least absolute deviations problems. Again, semimetric properties can be imposed by linear constraints.

## 5 Conclusions and Future Work

We proposed a new embedding method which enables: efficient classification and regression of functions of single vectors, efficient approximation of distance functions, and general, non-Euclidean, semimetric learning. To the best of our knowledge, this is the first work that enables learning arbitrary semimetrics. That is, our method is a universal semimetric learning and approximation method that can approximate any distance measure with as high accuracy as needed with or without semimetric constraints.

Interesting future work includes evaluating the method empirically on real-world datasets, both for single vectors classification and regression and for metric learning applications. It would also be interesting to develop efficient clustering and nearest neighbor algorithms for ID. The project homepage including code is at: http://www.ariel.ac.il/sites/ofirpele/ID/

## References

[1] K. Weinberger and L. Saul, "Distance metric learning for large margin nearest neighbor classification," *JMLR*, 2009.

[2] J. Davis, B. Kulis, P. Jain, S. Sra, and I. Dhillon, "Information-theoretic metric learning," in *ICML*, 2007.

[3] M. Schultz and T. Joachims, "Learning a distance metric from relative comparisons," in *NIPS*.

[4] E. Xing, A. Ng, M. Jordan, and S. Russell, "Distance metric learning with application to clustering with side-information," *NIPS*, 2003.

[5] A. Bar-Hillel, T. Hertz, N. Shental, and D. Weinshall, "Learning distance functions using equivalence relations," in *ICML*, 2003.

[6] J. Goldberger, S. Roweis, G. Hinton, and R. Salakhutdinov, "Neighbourhood components analysis," *NIPS*, 2005.

[7] A. Globerson and S. Roweis, "Metric learning by collapsing classes," *NIPS*, 2006.

[8] D. Ramanan and S. Baker, "Local distance functions: A taxonomy, new algorithms, and an evaluation," *PAMI*, 2010.

[9] B. McFee and G. Lanckriet, "Learning multi-modal similarity," *JMLR*, 2011.

[10] L. Yang and R. Jin, "Distance metric learning: A comprehensive survey," *MSU*, 2006.

[11] B. Kulis, "Metric learning: A survey," *FTML*, 2012.

[12] A. Bellet, A. Habrard, and M. Sebban, "A survey on metric learning for feature vectors and structured data," *arXiv*, 2013.

[13] S. Chopra, R. Hadsell, and Y. LeCun, "Learning a similarity metric discriminatively, with application to face verification," in *CVPR*, 2005.

[14] M. Cuturi and D. Avis, "Ground metric learning," *JMLR*, 2014.

[15] F. Wang and L. J. Guibas, "Supervised earth mover's distance learning and its computer vision applications," in *ECCV*, 2012.

[16] D. Kedem, S. Tyree, F. Sha, G. R. Lanckriet, and K. Q. Weinberger, "Non-linear metric learning," in *NIPS*, 2012.

[17] W. Yang, L. Xu, X. Chen, F. Zheng, and Y. Liu, "Chi-squared distance metric learning for histogram data," *MPE*, 2015.

[18] D. Ramanan and S. Baker, "Local distance functions: A taxonomy, new algorithms, and an evaluation," *PAMI*, 2011.

[19] S. Hauberg, O. Freifeld, and M. J. Black, "A geometric take on metric learning," in *NIPS*, 2012.

[20] O. Pele, B. Taskar, A. Globerson, and M. Werman, "The pairwise piecewise-linear embedding for efficient non-linear classification," in *ICML*, 2013.

[21] S. Maji, A. Berg, and M. J., "Efficient classification for additive kernel SVMs," *PAMI*, 2012.

[22] A. Rahimi and B. Recht, "Random features for large-scale kernel machines," *NIPS*, 2007.

[23] A. Vedaldi and A. Zisserman, "Efficient additive kernels via explicit feature maps," *PAMI*, 2012.

[24] F. Perronnin, J. Senchez, *et al.*, "Large-scale image categorization with explicit data embedding," in *CVPR*, 2010.

[25] Y. Chang, C. Hsieh, K. Chang, M. Ringgaard, and C. Lin, "Training and testing low-degree polynomial data mappings via linear SVM," *JMLR*, 2010.

[26] O. Pele, A. Globerson, and M. Werman, "Interpolated-discretized metric learning using linear programming," in *Distance Functions: Theory, Algorithms and Applications*, ch. 5, The Hebrew University of Jerusalem, 2011.

[27] A. Bernal, K. Crammer, and F. Pereira, "Automated gene-model curation using global discriminative learning," *Bioinformatics*, 2012.

[28] "Hypercube triangulation." https://www.ics.uci.edu/~eppstein/junkyard/cube-triangulation.html.

[29] Y. Rubner, C. Tomasi, and L. J. Guibas, "The earth mover's distance as a metric for image retrieval.," *IJCV*, 2000.

[30] O. Pele and M. Werman, "A linear time histogram metric for improved sift matching," in *ECCV*, 2008.

[31] O. Pele and M. Werman, "Fast and robust earth mover's distances," in *ICCV*, 2009.

[32] P. Indyk and N. Thaper, "Fast image retrieval via embeddings," in *IWSCTV*, 2003.

[33] Q. Lv, M. Charikar, and K. Li, "Image similarity search with compact data structures," in *CIKM*, 2004.

[34] K. Grauman and T. Darrell, "The pyramid match kernel: Efficient learning with sets of features," *JMLR*, 2007.

[35] S. Shirdhonkar and D. Jacobs, "Approximate earth mover's distance in linear time," in *CVPR*, 2008.

[36] S. Khot and A. Naor, "Nonembeddability theorems via Fourier analysis," *MA*, 2006.

[37] M. Perrot, A. Habrard, D. Muselet, and M. Sebban, "Modeling perceptual color differences by local metric learning," in *ECCV*, 2014.